\documentclass[journal,twoside,web]{ieeecolor}

\usepackage{jsen}
\usepackage{cite}
\usepackage{amsmath,amssymb,amsfonts}
\usepackage{algorithmic}
\usepackage{graphicx}
\usepackage{textcomp}
\usepackage{wrapfig}
\usepackage{orcidlink}
\usepackage{siunitx} 
\usepackage{textcomp}
\usepackage{xcolor}
\usepackage{url}
\usepackage{svg}
\usepackage{tabularx}
\usepackage{booktabs}
\usepackage{makecell}
\usepackage[nolist]{acronym}
\usepackage{subcaption}
\usepackage{soul}
\usepackage{flushend}
\usepackage{multirow}

\usepackage{booktabs}
\newcommand{\tabitem}{~~\llap{\textbullet}~~}

\newcolumntype{C}{>{\centering\arraybackslash}X}

\usetikzlibrary{shapes,arrows,chains}

\def\BibTeX{{\rm B\kern-.05em{\sc i\kern-.025em b}\kern-.08em
    T\kern-.1667em\lower.7ex\hbox{E}\kern-.125emX}}

\definecolor{abstractbg}{rgb}{0.89804,0.94510,0.83137}
\begin{document}

\title{Evaluation of a Smart Mobile Robotic System for Industrial Plant Inspection and Supervision}
\author{
        Georg K.J. Fischer*\orcidlink{0000-0003-1460-1061},
        Max Bergau*\orcidlink{0009-0000-4068-8936},
		D. Adriana G\'omez-Rosal*\orcidlink{0000-0003-2373-9716},\\
        Andreas Wachaja\orcidlink{0009-0002-6843-7519},
        Johannes Gr\"ater\orcidlink{0009-0004-2166-6749},
        Matthias Odenweller,
        Uwe Piechottka\orcidlink{0009-0007-6375-5805},
        Fabian H\"oflinger\orcidlink{0000-0001-5877-1439},
        Nikhil Gosala\orcidlink{0000-0003-3226-5356},
        Niklas Wetzel\orcidlink{0000-0002-0502-3215},
        Daniel B\"uscher\orcidlink{0000-0001-9321-3029},
        Abhinav Valada\orcidlink{0000-0003-4710-3114},
		Wolfram Burgard\orcidlink{0000-0002-5680-6500}
\thanks{* Denotes equal contribution}
\thanks{\textit{Corresponding author: Georg K.J. Fischer.}}
\thanks{G.K.J. Fischer is with the Fraunhofer Institute for Highspeed Dynamics, Ernst-Mach-Institute (EMI), Freiburg, Germany (email: georg.fischer@emi.fraunhofer.de).}
\thanks{M. Bergau is with the Sensors Automation Lab, Endress+Hauser Digital Solutions GmbH, Freiburg, Germany.}
\thanks{D. A. G\'omez-Rosal, N. Gosala, N. Wetzel, D. B\"uscher and A. Valada are with the Department of Computer Science, University of Freiburg, Germany.}
\thanks{A. Wachaja and J. Gr\"ater are with dotscene GmbH, Freiburg, Germany.}
\thanks{M. Odenweller and U. Piechottka are with Evonik Operations GmbH, Essen, Germany.}
\thanks{F. H\"oflinger is with Telocate GmbH, Freiburg, Germany.}
\thanks{W. Burgard is with University of Technology Nuremberg, Germany.}
\thanks{This work was supported by the German Ministry of Education and Research (BMBF) under the grant FKZ~16ME0023K (``Intelligentes Sensorsystem zur autonomen Überwachung von Produktionsanlagen in der Industrie 4.0 - ISA4.0'').}
}

\IEEEtitleabstractindextext{%
\fcolorbox{abstractbg}{abstractbg}{%
\begin{minipage}{\textwidth}%
\begin{wrapfigure}[12]{r}{3in}%
\centering\includegraphics[width=2.4in]{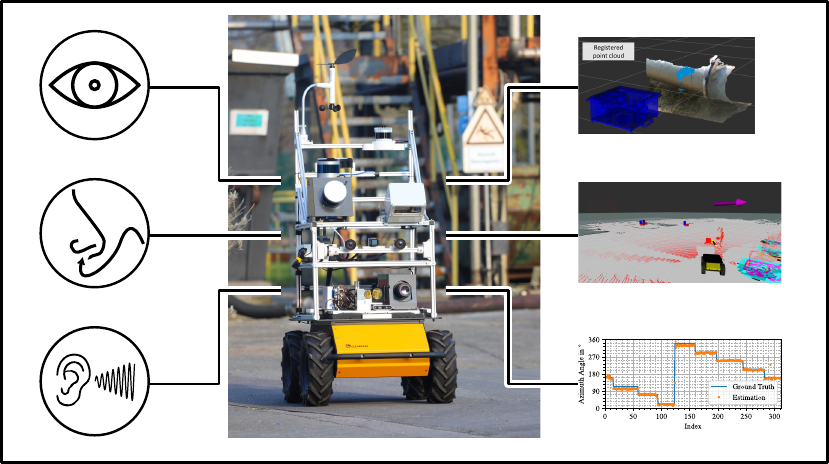}%
\end{wrapfigure}%
\begin{abstract}

Automated and autonomous industrial inspection is a longstanding research field, driven by the necessity to enhance safety and efficiency within industrial settings. In addressing this need, we introduce an autonomously navigating robotic system designed for comprehensive plant inspection. This innovative system comprises a robotic platform equipped with a diverse array of sensors integrated to facilitate the detection of various process and infrastructure parameters. These sensors encompass optical (LiDAR, Stereo, UV/IR/RGB cameras), olfactory (electronic nose), and acoustic (microphone array) capabilities, enabling the identification of factors such as methane leaks, flow rates, and infrastructural anomalies.
The proposed system underwent individual evaluation at a wastewater treatment site within a chemical plant, providing a practical and challenging environment for testing. The evaluation process encompassed key aspects such as object detection, 3D localization, and path planning. Furthermore, specific evaluations were conducted for optical methane leak detection and localization, as well as acoustic assessments focusing on pump equipment and gas leak localization.

\end{abstract}

\end{minipage}}}

\maketitle

\IEEEpeerreviewmaketitle

\begin{acronym}[JSONP]\itemsep0pt
    \acro{UCA}{Uniform Circular Array}
    \acro{DoA}{Direction-of-Arrival}
    \acro{MEMS}{Micro-Electro-Mechanical Systems}
    \acro{CDF}{Cumulative Distribution Function}
    \acro{SSL}{Sound Source Localization}
    \acro{HMI}{Human-Machine Interaction}
    \acro{CSS}{Coherent Subspace}
    \acro{FDFIB}{Frequency-Domain Frequency-Invariant Beamformer}
    \acro{SRP-PHAT}{Steered Response Power Phase Transform}
    \acro{MUSIC}{Multiple Signal Classification}
    \acro{ROS}{Robot Operating System}
    \acro{MOX}{Metal Oxide Semiconductor}
    \acro{AGV}{Automated Guided Vehicle}
    \acro{SBC}{Single Board Computer}
    \acro{LED}{Light-Emitting Diode}
    \acro{LiDAR}{Light Detection and Ranging}
    \acro{IMU}{Inertial Measurement Unit}
    \acro{OGI}{Optical Gas Imaging}
    \acro{IR}{Infrared}
    \acro{TDLAS}{Tunable Diode Laser Absorption Spectroscopy}
    \acro{DFB}{Distributed Feedback}
    \acro{FPGA}{Field Programmable Gate Array}   
    \acro{CNN}{Convolutional Neural Network}   
\end{acronym}

\section{Introduction}
Chemical production plants must adhere to stringent safety standards, and the responsibility of routine plant inspection typically falls on trained professionals. These inspections involve recording essential process parameters and identifying anomalies during safety rounds. This may include assessing local gauge values, visually inspecting for abnormalities, conducting olfactory and acoustic inspections to detect gas or fluid leaks, and identifying changes in the environment such as broken pipes or blocked passages.
In environments within Ex-Areas, or those that are harsh or remote, deploying human inspectors can pose significant dangers and incur financial burdens. Alternatively, the installation of fixed sensor equipment, while effective, can be costly and add to the maintenance workload. Moreover, the deployment of high-precision sensors across an entire plant is often economically impractical, especially when looking at older brownfield plants. Recognizing these challenges, there is a growing interest in mobile inspection robots.
Recent advancements in robotic technology, exemplified by robots such as Boston Dynamics Spot and others, have made it possible to access more challenging areas of industrial plants ~\cite{Guizzo.2019}. While these robots primarily serve as equipment carriers, numerous questions remain unanswered. These include determining the optimal sensors for deployment, devising effective data processing methods, and developing efficient navigation strategies for these robots within the plant environment. As mobile inspection robots gain prominence, addressing these questions becomes imperative for their successful integration into industrial safety and inspection processes.

Sensing the state of industrial processes entails a diverse array of sensors. For instance, detecting and localizing gas leakages in chemical plants involves a range of sensors, from metal oxide sensors to laser-based detection systems \cite{Francis.2022}. This diversity extends across industries, addressing changes in the environment such as obstacles obstructing critical transport routes, malfunctioning equipment (e.g., pipes, motors, valves), and parameters such as foam build-up that require optical assessment.
Replacing human inspection professionals presents a challenging task, as each use case typically requires bespoke development. This may involve the definition of specific physical parameters to observe, selecting an appropriate sensor, establishing a tailored signal, and data processing chain. However, different use cases can be independently implemented, with the robot serving the role of safely transporting measurement equipment to and from sites.
Advancements in mapping~\cite{vodisch2022continual, arce2023padloc}, localization~\cite{sorrenti2020cmrnet}, and navigation~\cite{younes2023catch} in mobile robotics are essential for achieving human-level navigation autonomously, encompassing tasks such as object detection and obstacle avoidance.
Although the opportunities for autonomous supervising robots in industrial plants are growing, few works of robots equipped with different sensing capabilities for such tasks have been reported or are still at a development stage, and furthermore, fewer works have been validated in a fully working industrial plant. This work aims to contribute to this field.
In this paper, we introduce a system that outlines the evolutionary trajectory in this domain. We delve into the development of the robot's navigation capabilities and present a subset of developed use cases focused on gathering essential process information.

This paper serves as an extension of the previously presented conference paper\cite{GomezRosal.2023}, augmenting the results with tangible data and providing an extensive literature review of existing works. Additionally, this contribution includes a more comprehensive exploration of the working principles of our inspection system and the associated sensors in a dedicated section.

\section{Related Work}

Robots have established their presence in the manufacturing sector, particularly in roles involving goods transportation within factories and warehouses, as seen with \acp{AGV}\cite{Zanchettin.2016, Oyekanlu.2020}. However, the application of mobile robotics in plant inspection presents a more complex challenge. On one hand, specific sensors need to be developed for particular use cases, while on the other hand, navigation and locomotion capabilities must reach a level that ensures safe and secure operation in hazardous areas and effective collaboration with humans and other vehicles.

\noindent\textbf{Robotic System:} 
The concept of industrial inspection robots is not new; nevertheless, mobile robots for industrial inspection is an active research topic. Furthermore, their mobility within dynamic environments is a more recent development \cite{M.Edwards.1984,Shneier.2015,Duran.2002,Karkoub.2021,Zhang.2022b,Dobie.2011}. Early works addressed inspection of industrial pipelines \cite{review_pipe_robots} and nuclear power plants\cite{robot_nuclear_plant}, \cite{robot_nuclear_plant2006}. Later, these proposals evolved to include a broader range of industrial applications, such as inspection of waste deposits and electric power cables\cite{underground_cable_systems}. Soon after, these experiments ventured into the civil realm with inspection of building facades, concrete constructions\cite{robot_Construction_Automation}, structures such as ships, wind turbines, and aircraft\cite{almadhoun2016survey}, and even examination of underwater facilities\cite{underwater_inspection}. 
Recent work in plant inspection\cite{Ishida.2012} reviews chemical sensing applications for mobile robots, addressing challenges such as the quest for improved chemical sensors and the development of more intelligent behavior.

In the context of diverse industries, a study \cite {Yu.2019} investigates various approaches to inspection in the oil and gas sector. This review encompasses diverse designs for inspections, including those tailored for vertical structures, pipelines, and underwater environments. The introduction of robots into inspection processes significantly enhances safety by eliminating the need to deploy humans in risky areas. Yet, expertise in technology deployment is crucial, as highlighted in another study \cite{Kas.2020} that delves into various use cases for robots in hazardous sites, emphasizing the need for specialized knowledge in deploying this technology.

Focusing on the exploration and mapping of gas sources, a specific study \cite{Francis.2022} is centered on gas detection principles and reviews existing works proposing mobile inspection vehicles. Together, these works showcase the evolving landscape of mobile robotics in-plant inspection, emphasizing both the technological advancements and the need for domain-specific expertise in deploying these systems effectively and safely.\\
However, autonomously conducted inspections of industrial settings, such as processing plants, present ongoing research challenges, such as regarding autonomous navigation, retrieval of measurements in disordered environments, or its consequent interpretation. Hence, to faithfully relieve human inspectors from these challenging tasks, the proposals need to enhance their intelligence and reliability. To this regard, few initiatives aim to design an inspection robot with all their required functions as a centralized, and yet flexible, system. This is the gap that our system seeks to address. \\
\noindent\textbf{Microphone Array:} \ac{SSL} is a prevalent subject in robotics, with a comprehensive review of common methods provided in \cite{CalebRascon.2017}. Its applications span \ac{HMI}, speaker or asset localization, and event detection such as identifying leakages in the surrounding environment \cite{Nakamura.2012,Fischer.2022,J.M.Valin.2003,Schenck.2019}. While gas or air leak localization is crucial for industrial inspection and has been previously addressed, existing approaches often rely on fixed installed acoustic transducers and static time-domain features for locating points of leakage in pipes \cite{FangWang.2017}. The work in \cite{Yan.2018} applies a four element linear array to find two leakage sources in the emission range of \SIrange{63}{187}{\kilo\hertz}.
The work in \cite{Schenck.2019} utilizes ultrasonic emissions from leaks in pressurized pipes, employing a peak search on a beam-formed spectrum. By leveraging multiple poses of a robot, potential leaks are localized, necessitating a microphone array with 32 elements. In contrast, our work introduces an algorithm capable of achieving accurate sound source localization with a significantly reduced number of microphones, providing a more resource-efficient solution for industrial applications.\\
\noindent\textbf{Gas Cameras:} Modern \ac{OGI} cameras typically utilize a mid- or longwave-infrared camera in combination with a bandpass filter to detect and visualize gas leaks \cite{ravikumar_are_nodate}. The gas of interest absorbs parts of the omnipresent background infrared radiation, leading to a contrast in the \ac{IR} camera image. This principle allows for the detection of a wide range of gases, including methane, ammonia, and sulfur hexafluoride. In this work, we present an active gas camera that does not rely on background illumination. Instead, it employs its own active illumination in the form of a laser which allows for more sensitivity, gas selectivity as well as accurate concentration estimation. It comes to the cost of a shorter detection range and the need of a background \cite{Bergau.2023}. Such active approaches are yet rare due to costs and complexity. A laboratory setup has been published by  Strahl et al. \cite{strahl_methane_2021} and a much simplified version of a camera by Nutt et al. \cite{nutt_developing_2020}. Commercially, the company QLM Technologies is working on an advanced active scanning approach \cite{titchener_single_2022}. 


The main contributions of this paper are:
\begin{itemize}
    \item An autonomous mobile robotic system equipped with a diverse and integrated set of sensors, including olfactory, optical, and acoustic capabilities. This advanced system is designed not only for gas detection but also for infrastructure monitoring. 
    \item A robot navigation framework, enabling flexible and independent mobility within industrial settings.
    \item In-depth examination of the performance of the integrated sensors under diverse conditions. Notably, the study includes a detailed analysis of sensor functionality within the wastewater treatment facility of a chemical plant situated in Marl, Germany. 
    \item To the best of the authors' knowledge, an active gas camera has been mounted on a mobile robotic platform for the first time. Since the active approach requires reflected light from a background, the robot has the potential to adapt its filming position, thereby finding ideal illumination locations.
    \item Apart from hyperspectral imaging which is typically significantly more expensive, to the best of our knowledge for the first time methane leaks as small as \SI{40}{\milli\liter\per\minute} leaks have been filmed in industrial conditions in real-time.

\end{itemize}

\section{System Architecture}

\begin{figure}
    \centering
    \resizebox{\linewidth}{!}{\input{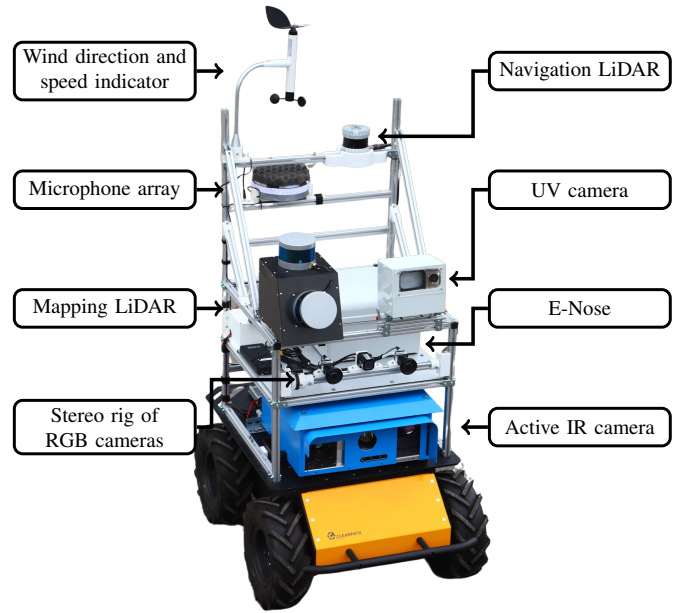}}
    \caption{The mobile robotic platform with multimodal sensors.}
    \label{fig:robot-annotated}
\end{figure}

Our robotic system, shown in \autoref{fig:robot-annotated}, consists of a mobile robot platform integrated with diverse multimodal sensors, summarized in \autoref{tab:overview-sensors}. 
The robot navigates the industrial plant autonomously with its onboard processing capabilities.
However, tasks that require more resources, such as object or anomaly detection, are executed on a remote server.
To this end, the data is transmitted over the wireless network employing the \ac{ROS}~\cite{ROS}. \autoref{tab:overview-sensors} gives an overview of the equipped sensors and devices.

\begin{table*}
    \centering
    \caption{Overview of equipped sensors and devices}
    \renewcommand{\arraystretch}{1.5}%
    \begin{tabularx}{\linewidth}{lll}
    \toprule
        \textbf{Device} & \textbf{Specs.} & \textbf{Use Cases }\\ 
        \midrule
        \multirow{1}{*}{Electronic Nose} & \tabitem 3~$\times$~non-selective MOX sensors & \tabitem Methane, CO2, flammable gas detection \\
        \midrule
        \multirow{3}{*}{Active IR Camera} & \tabitem Methane rate $\geq$\SI{40}{\milli\liter\per\minute}                                        & \tabitem Methane Leak Localization \\
                                          & \tabitem Wind speed $\leq$\SI{2}{\meter\per\second} &  \tabitem Concentration Length Measurements \\
                                          &                                                             \tabitem Distance to Leak $\leq$\SI{3}{\meter}          & \tabitem Flowrate Quantification \\
        \midrule
        \multirow{2}{*}{UV Camera} & \tabitem \SI{365}{\nano\meter} UV-LEDs & \tabitem UV-excited fluorescence \\
                                   & \tabitem OpenMV Cam H7R2 CMOS camera   & \tabitem Oil film detection \\
        \midrule
        \multirow{3}{*}{Microphone Array} & \tabitem \ac{UCA}                                  & \tabitem Acoustic anomaly detection \\
                                          & \tabitem 5~$\times$~TDK ICS-40720 MEMS-microphones & \tabitem Acoustic machine condition monitoring \\
                                          & \tabitem $\varnothing $ \SI{6.8}{\centi\meter}     & \tabitem Leakage detection \\
        \midrule
        \multirow{2}{*}{Mapping \ac{LiDAR}} & \tabitem 2~$\times$~Velodyne VLP16 laser scanner & \tabitem 3D mapping                                 \\
                                       & \tabitem IMU                                     & \tabitem Geometric changes and anomaly detection \\
        \midrule
        \multirow{2}{*}{Passive Cameras} & \tabitem 2$\times$ RGB Blackfly S GigE FLIR cameras           & \tabitem Dynamic object detection (pedestrians, obstacles, etc.) \\
                                         & \tabitem FLIR Boson Long wave infrared (LWIR) thermal camera  &                                                          \\
        \midrule
        \multirow{2}{*}{Navigation LiDAR} & \tabitem OS128 Ouster LiDAR.                      & \tabitem 2D mapping \\
                                          & \tabitem \SI{90}{\deg} vertical field of view, \SI{50}{\meter} range                        
                                                  & \tabitem Localization and navigation \\
        \midrule
    \bottomrule
    \end{tabularx}
    
    \label{tab:overview-sensors}
\end{table*}

\begin{figure*}
    \centering
    \resizebox{\linewidth}{!}{\includegraphics[width=1.0\linewidth]{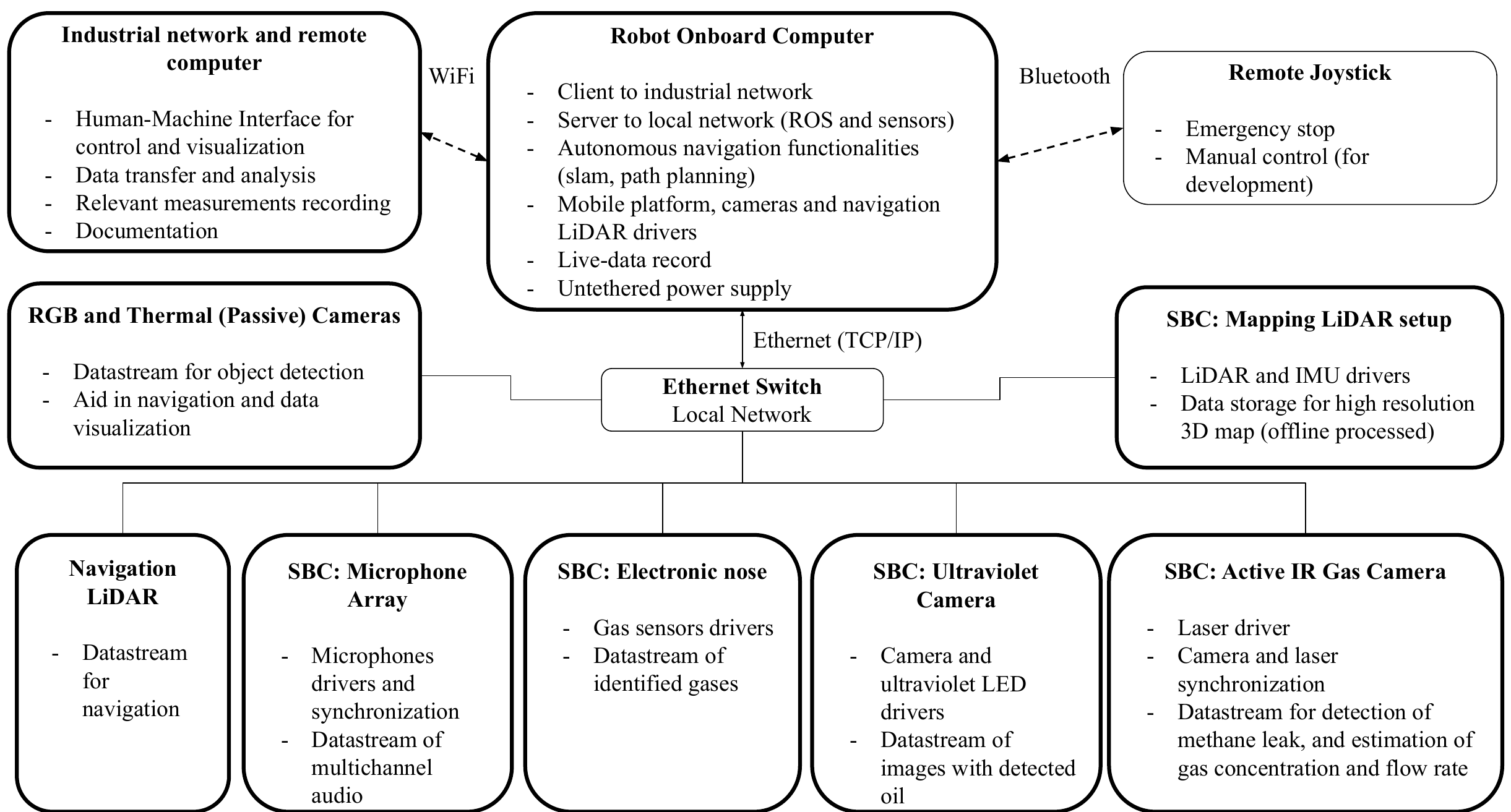}}

    \caption{System architecture with communication between the components. Solid or dotted lines between the boxes represent, respectively, Ethernet or wireless connectivity. 
    }
    \label{fig:system-diagram}
\end{figure*}

The core software component responsible for centralizing the interactions, control, and data access is deployed on the robot's onboard computer and is managed by the \ac{ROS} middleware. It provides modules for package management, hardware abstraction, and communication in a networked fashion. This enabled the connectivity between the robot and desktop computers to each \ac{SBC} contained in the sensing devices.

Figure \ref{fig:system-diagram} shows the system architecture and communication channels, comprised of the computer cluster on the mobile robot platform and the stationary components, such as the industrial remote computer and the remote joystick control. Most of the sensing devices count with a \ac{SBC}, which is responsible for the connectivity to other onboard computers via a Local Ethernet Network, among other data configuration and processing functions. 

The system is controlled live from a computer through the industrial WLAN, and for development purposes, the robot platform is manually controlled with a remote joystick. The onboard sensing devices stream the data via the ROS middleware; where predefined data types are used for internal communication.

\subsection{Electronic Nose}

For the task of gas anomaly detection, we developed an \textit{Electronic Nose} (E-Nose).
Its working principle is based on the integration of different gas sensors.
We employ three non-selective \ac{MOX} sensors and,
in combination with optical (non-dispersive infrared, laser scattering) technologies,
we measure air contaminants, methane, CO2, and flammable gases.
The second category of gas sensors, based on electrochemical principles, are specific and fast in their response.
This data allowed the training of a supervised learning network able to identify gas signatures.
Additionally, relevant information, such as humidity, wind speed, and direction, was logged.

\subsection{Active Infrared Gas Camera}

We developed an active infrared gas camera (IRcam) for the sensitive detection of methane leaks, and the estimation of gas concentration and flow rate.
The light detection is carried out by an ImageIR 8300 camera from Infratec,
while its field of view is illuminated by a \SI{25}{\milli\watt} tunable $\sim$\SI{3260}{\nano\meter} single mode \ac{DFB}-interband cascade laser from nanoplus. 
\begin{figure}
    \centering
    \includegraphics[width=1.0\linewidth]{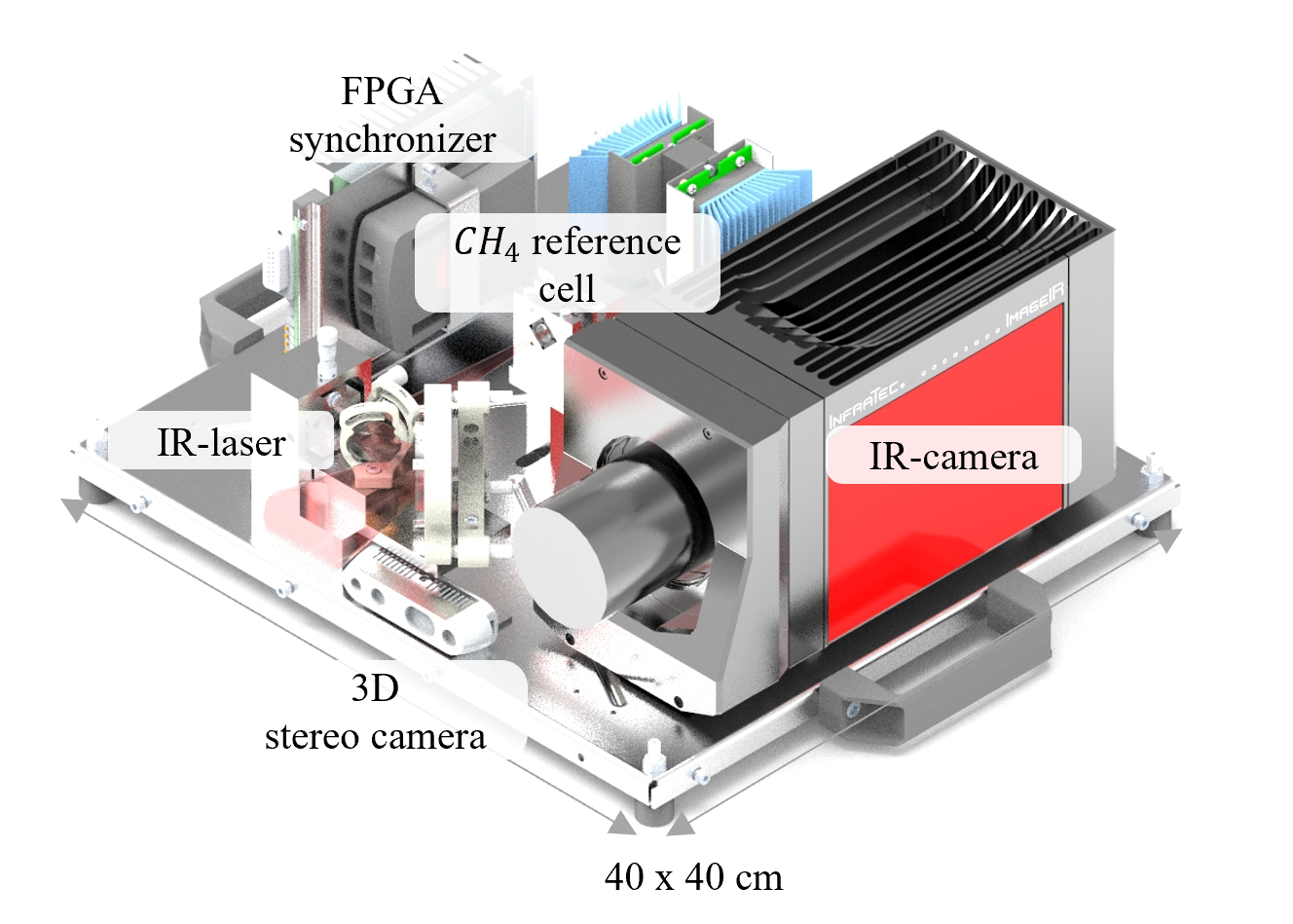}
    \caption{Setup of the active gas camera. A detailed description of its working principle can be found in \cite{bergau_real-time_2023}.}
    \label{fig:ir_cam}
\end{figure}

Three consecutive images are recorded while the laser's wavelength is tuned over a methane absorption line.
We process the image batches as described in~\cite{bergau_real-time_2023} using \ac{TDLAS} to obtain concentration length information. Thereby, the synchronization between laser tuning and camera trigger is achieved with a redpitaya STEMlab 125-10 single board computer with an \ac{FPGA} on board.
To ensure wavelength stability across different temperatures as well as to handle the aging effects of the laser diode, a reference gas cell is filled with methane. Less than 1\% of the laser light is coupled out of the diverging laser beam and fed through the cell as shown in figure \ref{fig:ir_cam}.  
We feed the gas-image stream at \SI{5}{\hertz} to a 3D \ac{CNN} trained to estimate the flow rate of the gas leak in real time. Due to the nature of the \ac{TDLAS} based gas detection, we gain more precise concentration length information than state-of-the-art gas cameras, which leads to more accurate flow rate predictions \cite{bergau_flow_2023, wang_videogasnet_2022}.

\subsection{Ultraviolet Camera}

We developed a remote fluorescence detection system for the recognition of oil films on surfaces and potential oil rests.
This system consists of a \SI{365}{\nm} Ultraviolet (UV) Light-emitting Diode (LED), ODS75 Smart Vision Lights, and a CMOS camera, OpenMV Cam H7R2. 
Pairs of images are taken by the camera, one image with ambient light and one with additional UV illumination.
The difference between these two frames yields an image containing UV-excited fluorescence,
to reveal information about oil films due to their aromatic molecular structures~\cite{stasiuk97}.
While this sensor is able to detect oil films up to a distance of \SI{1}{\m} in indoor lighting conditions,
the sensitivity decreases outdoors, in particular in sunny conditions.

\subsection{Microphone Array}
\begin{figure}
    \centering
    \includegraphics[width=1\linewidth]{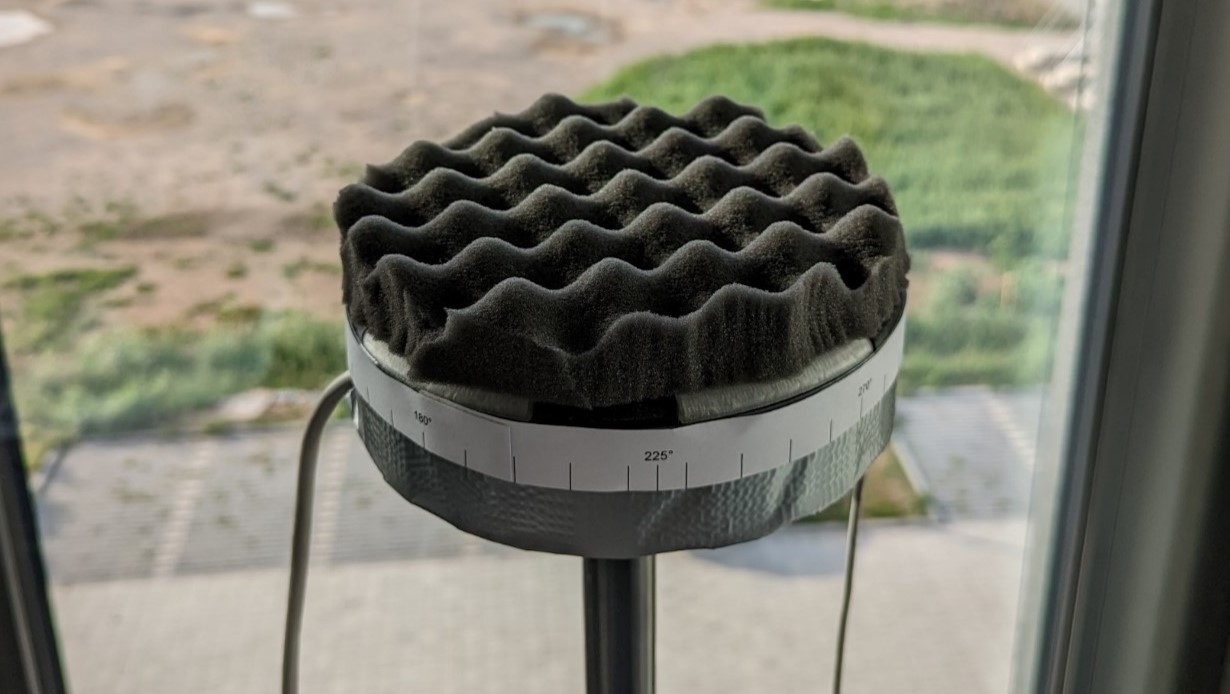}
    \caption{\ac{UCA} Microphone Array with a windscreen for outdoor usage. Details of the employed algorithms can be found in \cite{Fischer.2021}}
    \label{fig:microphone-array-photo}
\end{figure}

To provide the robot with hearing capabilities, we developed a five-microphone uniform circular array as shown in \autoref{fig:microphone-array-photo}.
We employ TDK ICS-40720 \ac{MEMS} microphones with a sampling rate of \SI{96}{\kilo\hertz},
and an inter-microphone spacing of $\approx\, $\SI{8}{\centi\meter}.
The setup is capable of localizing sound sources with a maximum frequency of around \SI{2.1}{\kilo\hertz}
using a specifically designed method for \ac{DoA} estimation. 
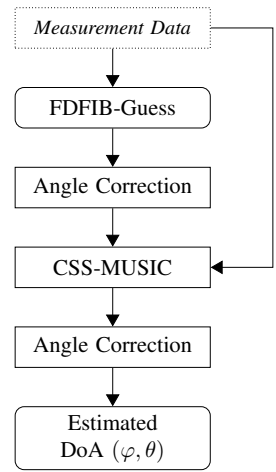
\begin{figure}
    \centering
    \resizebox{.4\linewidth}{!}{\begin{tikzpicture}[%
    >=triangle 60,              
    start chain=going below,    
    node distance=6mm and 60mm, 
    every join/.style={norm},   
    ]
    \tikzset{
      base/.style={draw, on chain, on grid, align=center, minimum height=4ex},
      proc/.style={base, rectangle, text width=8em},
      test/.style={base, diamond, aspect=2, text width=5em},
      term/.style={proc, rounded corners},
      coord/.style={coordinate, on chain, on grid, node distance=6mm and 25mm},
      nmark/.style={draw, cyan, circle, font={\sffamily\bfseries}},
      norm/.style={->, draw},
      free/.style={->, draw},
      cong/.style={->, draw},
      it/.style={font={\small\itshape}}
    }
    \node [proc, densely dotted, it] (p0) {Measurement Data};
    \node [term, join]      {FDFIB-Guess};
    \node [proc, join] (p1) {Angle Correction};
    \node [proc, join] (p2) {CSS-MUSIC};
    \node [proc, join]      {Angle Correction};
    \node [term, join] (t1) {Estimated DoA $\left(\varphi, \theta\right)$};

    \node [coord, right=of p1] (c2)  {};
    \path (p0.east) to node [near start, xshift=1em] {} (c2);
    \draw [->] (p0.east) -| (c2) |- (p2.east);		
\end{tikzpicture}}
    \caption{Sequential steps of the DoA estimation algorithm, starting with a coarse but computationally efficient guess and progressing to a more detailed peak search in the wideband MUSIC spectrum for refined accuracy.}
    \label{fig:microphone-array-doa-algo-flow}
\end{figure}

The algorithms under consideration have been thoroughly examined in a laboratory setting, as outlined in \cite{Fischer.2021}. These algorithms, originating from classical subspace \ac{DoA} estimation techniques such as \ac{MUSIC}, have been expanded to accommodate the \ac{UCA} configuration.
Recognizing that subspace methods are traditionally formulated for narrowband estimation, whereas wideband signals are expected (as detailed in a previous publication), these algorithms have been further adapted to support an arbitrary frequency range, extending up to the spatial Nyquist frequency.
The resulting algorithms comprise a two-step process, as depicted in \autoref{fig:microphone-array-doa-algo-flow}. Initially, a computationally inexpensive coarse \ac{FDFIB} guess step \cite{DoHong.2003} is implemented, followed by a more resource-intensive fine-search step within the \ac{CSS} \ac{MUSIC} spectrum \cite{H.Wang.1985}. This hybrid approach ensures an optimal balance between computational efficiency and accuracy.
Notably, the algorithms exhibit the capability to resolve multiple sources, facilitated by a selection criterion applied to the singular values of the spectral covariance matrix. However, it's important to note that the algorithm's performance is constrained by the number of available microphones in the array.

We detect sound anomalies by recognizing sound samples as normal or abnormal through a supervised learning model.
The network is trained on the MIMII~\cite{purohit2019mimii} dataset
and can be extended to include sound samples from the plant where the robot is operated.
The anomalies, constituted by unknown sound samples, are detected using a deep neural network with an encoder-decoder architecture.
To this end, the network learns features of the sound samples by encoding the sample into a latent vector embedding
and then, decoding the vector to recreate the input sound sample.
For unknown sounds, the input cannot be reconstructed, which indicates an anomaly.

\subsection{LiDAR Mapping System}
\label{sec:mapping-lidar}
We developed a task-specific \ac{LiDAR} setup  to build  accurate 3D maps of the environment of the robot. We furthermore map the deep sewage channels at the plant.
The unit comprises two Velodyne VLP16 laser scanners mounted on a cuboid structure.
While the top laser is leveled with the ground to capture the distant surroundings,
the front laser scanner is steeply inclined, allowing to capture the ground and possible channels in front of the robot.
We employ an \ac{IMU} to track the short-term motion of the system, increasing mapping accuracy and robustness.
A dense 3D map of the environment is obtained by fusing the continuously captured data with a 3D 
Simultaneous Localization and Mapping (SLAM) approach when the robot is in motion.
The resulting map of the environment is represented as a point cloud in 3D space.
With this generic representation, various applications are realized, such as embedding other sensor information (e.g., temperature),
geometric alignment for inventory purposes, changes or anomalies detection, and lastly semantic segmentation with deep learning techniques.

\subsection{Passive Cameras}

The optical perception system includes three forward-facing cameras composed of
a stereo rig of two RGB cameras, Blackfly S GigE FLIR, with a resolution of $1544\times2064$, 
and a thermal camera, FLIR Boson Long wave infrared (LWIR), with $512\times640$ pixels
and a spectral range of \SI{7.5}{\micro\meter}-\SI{13.5}{\micro\meter}.
These three cameras are synchronized with an external trigger at a frame rate of 30 FPS. 

The live object detection module employs the images from the RGB cameras to support the navigation functions. It detects dynamic agents in the scene, such as cars and pedestrians, and is based on CenterNet~\cite{centernet}, a 
network for object detection. CenterNet is a fully-supervised model that first predicts the object centers and their corresponding center offsets, and then combines these values to estimate bounding boxes around dynamic objects.

\subsection{Mobile Robot Navigation System}

We employ the Clearpath Husky A200 as the mobile robotic platform since it is a rugged, all-terrain robot equipped with an onboard computer that is designed to operate in indoor and outdoor environments. The robot achieves a safe maximum speed of \SI[per-mode = symbol]{1}{\meter\per\second} in forward movement with the sensor arrangement weighing about \SI{35}{\kilo\gram}. It also supplies electrical power to most of the sensors, enabling an untethered performance in a self-contained fashion. The onboard computer gets connected to the wireless industrial network, from which, a remote computer access and controls the system through the Human-Machine Interface over \ac{ROS}. 

To enable autonomous navigation, we use an OS128 Ouster LiDAR. 
It has a vertical field of view of 90\textdegree{}, \SI{50}{\meter} range, 128 times 1024 channels at a \SI{20}{\hertz} rate, and a built-in IMU.
The navigation functionality consists of the tasks of localization, mapping, and path planning.
While the localization approach in a known 2D grid map builds on Adaptive Monte Carlo Localization (AMCL)~\cite{probabilisticrobotics},
using a Rao-Blackwellized particle filter (RBPF), we solved the SLAM task with an efficient RBPF that creates grid maps~\cite{OpenslamGmapping}.
To this end, the 3D point clouds from the Ouster LiDAR were projected into 2D range scans.

The path planning task involves two hierarchized planners. While we solve the global planning using the $A^*$~\cite{Astar} approach, we employ the more advanced timed elastic bands (TEB)~\cite{TEB} approach for local planning. It is complemented by a reinforcement learning (RL) algorithm supporting the avoidance of dynamic obstacles. 
The RL algorithm uses the dynamic obstacles detected in the RGB images and projects them in the birds-eye view (BEV) using the Ouster LiDAR data.
We combine features extracted from this local semantic map via a CNN with the robot's low-dimensional internal state, such as the navigation goal and recent controls.

\section{Experiments}

The experiments were conducted at the wastewater treatment plant within the chemical facility of Marl, Germany. This industrial setting offered a ground-level route encompassing diverse points of interest, including tasks involving hot pipes, noisy pumps, and sewage canals with occasional variations in gas levels. The discussion in this section is divided into three main parts.
Firstly, the focus is on autonomous navigation capabilities. The presentation includes a detailed discussion of the experiments conducted to evaluate the robot's ability to navigate autonomously in this complex industrial environment.
Moving on to the second part, the section delves into gas sensing experiments. Special attention is given to the outcomes obtained from the infrared (IR) camera, providing a comprehensive analysis of the robot's performance in tasks related to gas detection within the plant.
In the final part of this section, the focus shifts to acoustic experiments. The initial set of experiments involves the localization of pump equipment using acoustic signals, followed by experiments on the detection and localization of acoustic leaks. The discussion concludes with the introduction of a more general approach to anomaly detection using acoustic data.

\subsection{Navigation Experiments}

Robot information, related to position and localization (Odometry), was computed through probabilistic approaches using the navigating LiDAR data. This point cloud was used in the tasks of Mapping, Localization, and Global/Local path generation through the Dijkstra algorithm and the Dynamic Window Approach, respectively. The 2D mapping capabilities were compared across resolutions from 0.01 to 0.2 m, which had a direct effect on the map size and update speed. 

\begin{figure*}
\centering
    \resizebox{\linewidth}{!}{\includegraphics[width=1.0\linewidth]{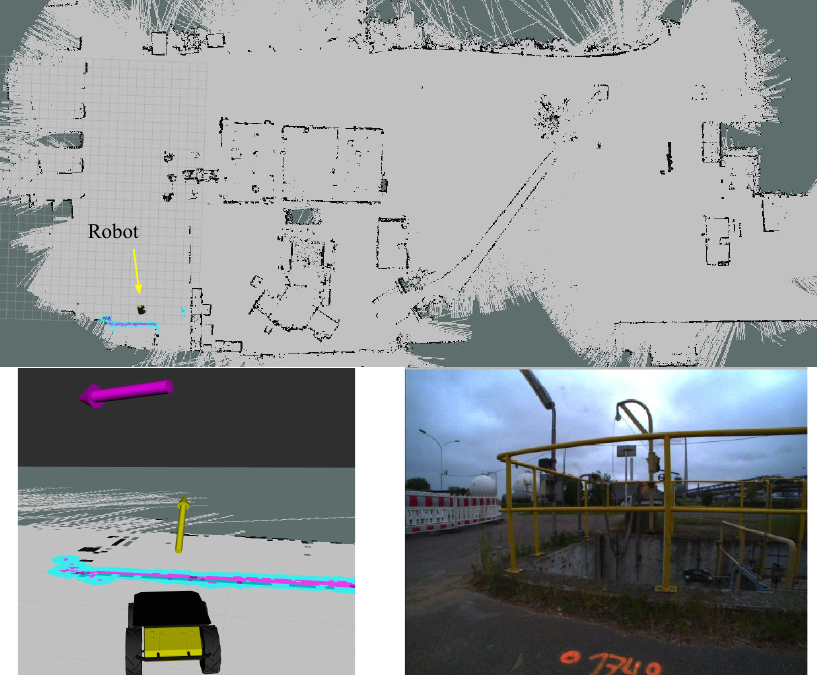}}

\caption{
Visualization of robot in map and markers for wind and audio direction. The three images depict the same instance of the robot navigating the test site in real life.
Upper: Complete 2D occupancy grid map of the test site, shown in gray and black from a top-down perspective. The position of the real robot is displayed on the map with its 3D model.
Lower left: Perspective rearview of the 3D model robot in the visualizer, in the same position as the upper figure. Its navigation in the map is enabled by the costmap (in cyan and magenta), which aligns with the map. Simultaneously, the purple arrow marks the wind direction (data obtained from the e-nose), and the yellow arrow signals the audio noise direction (data from the microphone-array). 
Lower right: Image from the RGB camera on the right side of the robot. The robot is front-facing a fence, which is recognized in the map and creates consequently the costmap. This supports the localization in the map.
} \label{fig:rvizMap}
\end{figure*}

The robot was tasked with navigation missions and a visualization interface from a remote station, allowed to know relevant live information from the robot and the sensors. This is displayed in Figure \ref{fig:rvizMap} with a representation of the real robot on a map alongside indicators sensor data. In the figure, the 2D occupancy grid map of the complete test site is displayed. When the "real" robot is navigating the site, its position is signed in the map by a 3D model of the robotic platform. As the real robot changes its position in the test site, the 3D model updates its localization through the costmap computed with the particle filter and the "live" LiDAR data. In the same Figure, the three images correspond to the same moment as the robot navigates the real-life test site. 
From the lower images, a closer look at the perspective rear view of the 3D model robot in the visualizer, shows the correspondence of the costmap with the 2D map, allowing a correct localization and consequently reliable navigation. Finally, an inspection of the "live" data transmitted by the RGB camera lets us know that the "real "robot is in front of a fence, acting as a landmark for the 2D map and enhancing the localization. Additionally, diverse RL agents were trained using the Soft Actor-Critic~\cite{softactorcritic} algorithm with varying hyperparameters, to compare the ability to navigate around dynamic objects in simulations and real-world scenarios predicatively.

To achieve more robust navigation in environments with dynamic obstacles, an object detection module for obstacle avoidance was developed. This module can recognize car and people categories, as well as determine their 2D and 3D extents. This leads to dynamic obstacle recognition and its avoidance while navigating. Such a task is divided into two subtasks: 1) recognition of specific objects in the scene using an object detection pipeline, and 2) projection of the 2D bounding boxes of the detected objects into the 3D world.

\paragraph{2D Object Detection}
The first subtask leverages the state-of-the-art object detection framework CenterNet, which is a single-stage object detection framework that accepts an RGB image as input and outputs a center point heatmap while simultaneously regressing the height, width, and offsets for every object in the image. These separate outputs are later fused with a post-processing step to generate the final 2D bounding boxes. To train CenterNet for the object recognition task, numerous labeled images were required in principle. To avoid the time-consuming task of manual image annotation for the object detection module training, we leveraged pre-existing object detection ground truths from a different domain and then adapted the model to work on the chemical plant dataset. To this end, we pre-trained the model using labels from the autonomous driving dataset KITTI~\cite{geiger2013vision} comprising 3712 train and 3769 validation samples and then fine-tuned it on 167 manually annotated images from our chemical plant and validating the model on 17 images. 
The evaluation of 20 manually annotated images resulted in Intersection over Union (IoU) values and Average Precision (AP) scores at the confidence threshold of 0.3 to 0.6855. These values were 0.300 and 0.668 for the Car class, and 0.300 and 0.703 for the Pedestrian class respectively. The accuracy of locating only the objects in the image without considering their width and height predictions is referred to as the AP of Centerpoint matches and obtained a value of 0.7662.


\paragraph{Estimation of 3D Position}
The second subtask concerns the estimation of the 3D locations of the detected objects and leverages the depth information obtained from the onboard LiDAR sensor along with the extrinsic calibration between the LiDAR and camera sensors. The LiDAR point cloud is projected onto the RGB image using the cameras' intrinsic parameters coupled with the extrinsic transformation between the two sensors. For every detected object, the LiDAR points outside the 2D extent of the bounding box are filtered out. Afterward, the bounding boxes are shrunk by a factor of 0.7 to alleviate potential noise introduced by incorrect intrinsic or extrinsic calibration matrices, thus improving the robustness of the model. 
The filtered LiDAR point cloud represents all the points belonging to the object being considered, which is then used to compute the 3D extent of the object. These 3D extents are then flattened along the height dimension to determine the extent of the object in the Bird’s Eye View (BEV) space. 

\begin{figure*}
\centering
    \resizebox{\linewidth}{!}{\includegraphics[width=1.0\linewidth]{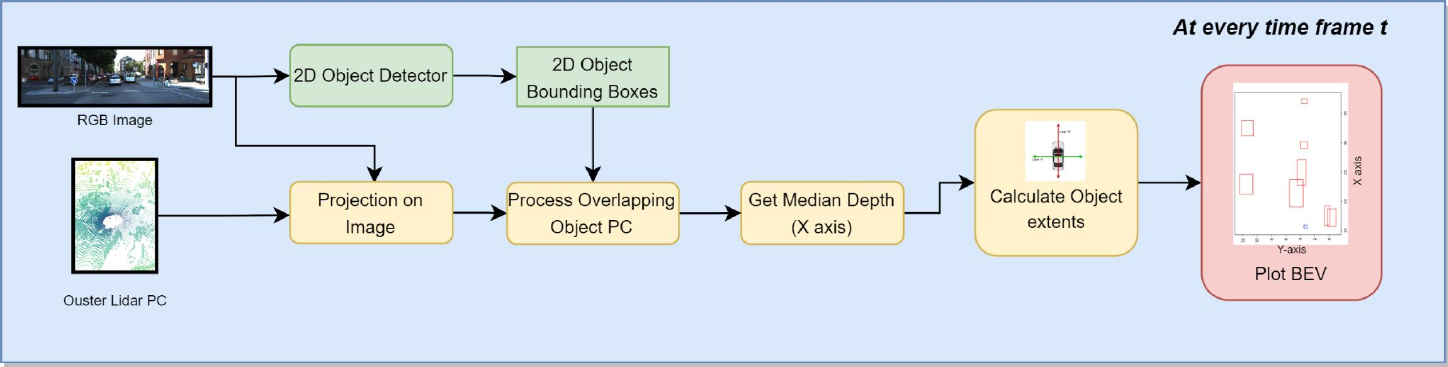}}
\caption{Pipeline of object detection and its 3D position estimation for robot navigation in a dynamic environment.  } \label{fig:objectdetection}
\end{figure*}

Figure ~\ref{fig:objectdetection} illustrates the pipeline of the object detection modules and its position estimation.
This information is utilized to plan safe collision-free paths around movable obstacles. 

\begin{figure*}
\centering
    \resizebox{\linewidth}{!}{\includegraphics[width=1.0\linewidth]{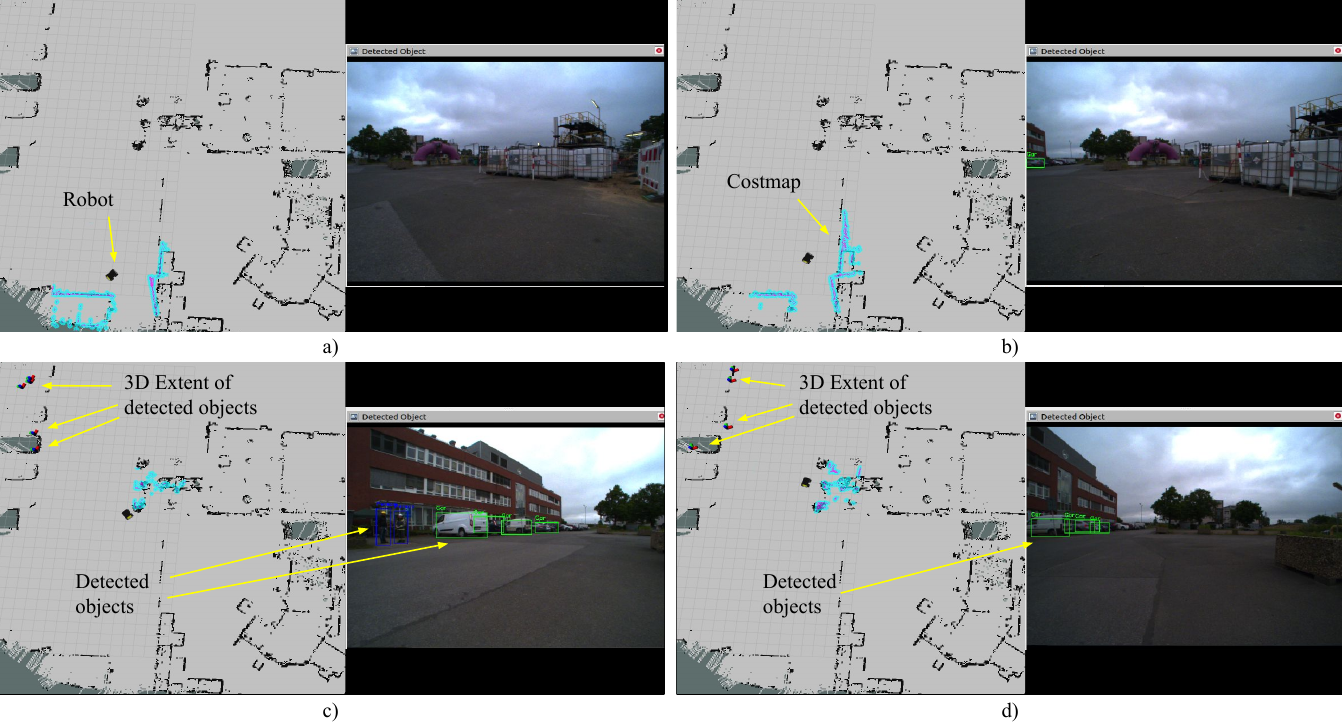}}
\caption{
Visualization from a top-view perspective of the robot navigating the test site and detecting objects. 
The order of images corresponds to successive time points following the alphabetical order, with 'a' being the initial moment, followed by 'b,' 'c,' and 'd.'
Every image displays the robot's real-world localization, featuring the robot model overlaid on the 2D map, its localization (costmap matching the map), and capture from the right RGB camera on the real robot. 
Upper left: First position of the robot in a northeast fashion. The accompanying RGB Image shows some white industrial containers, a white fence, and a purple pipe in the background. While the proximity of the containers and fence contributes to the robot's localization through the costmap, they remain unidentified by the object-detection module.
Upper right: Second position of the robot. Since the robot moved in the northeast direction, the costmap (colored in cyan) accompanies the movement and by matching the 2D map stabilizes the localization. The RGB image shows a closer look at the pink pipe, along with the first car detected in a tiny green box.
Lower left: Third position of the robot, where it moved in a northwest direction, losing the purple pipe from the field of view. The robot faces now a parking lot where people are also present. 
The detected objects are classified as human (in blue boxes) or as car (in green boxes) in the RGB Image and their 3D extents are projected in the 2D map with small markers.
Lower right: Final position of the robot, after a slight northeast movement from the previous position. Cars are still in the robot's field of view and therefore, their position in the 2D map is still displayed. The costmap matching the 2D map means that the robot is still properly localized. 
} \label{fig:rvizNav}
\end{figure*}

Figure ~\ref{fig:rvizNav} shows sequential captures of the robot navigating the test site in real life in alphabetical order. It is observed that as the robot navigates close to fixed objects at the time the map was created, the robot recognizes the features (such as fences, pipes, buildings, etc) and the generated costmaps enable the localization. Accordingly, these elements remain unidentified by the object detection module.
When a car or a human enters the field of view of the RGB cameras, the system identifies them and obtains their 3D extent. This data is consequently displayed in the 2D map with red and blue markers, and considered by the path planner to generate a collision-free path around these objects.

\subsection{3D Mapping and Change Detection}
\begin{figure}
\centering
\begin{subfigure}[c]{\linewidth}
    \includegraphics[width=\linewidth]{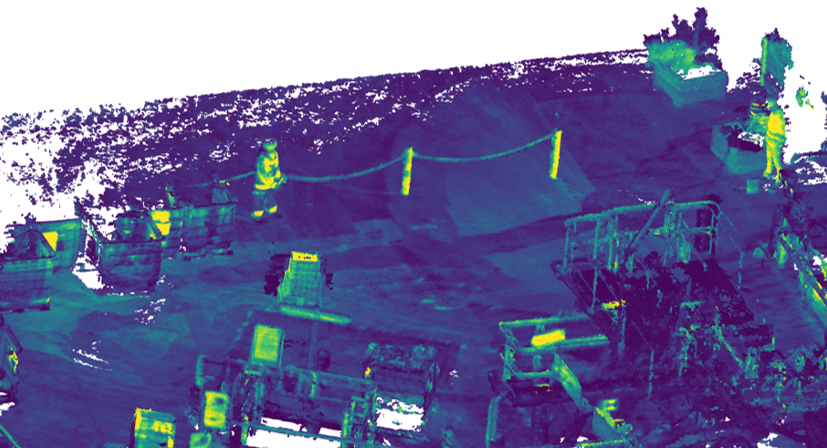}
\end{subfigure} %
\begin{subfigure}[c]{\linewidth}
    \includegraphics[width=\linewidth]{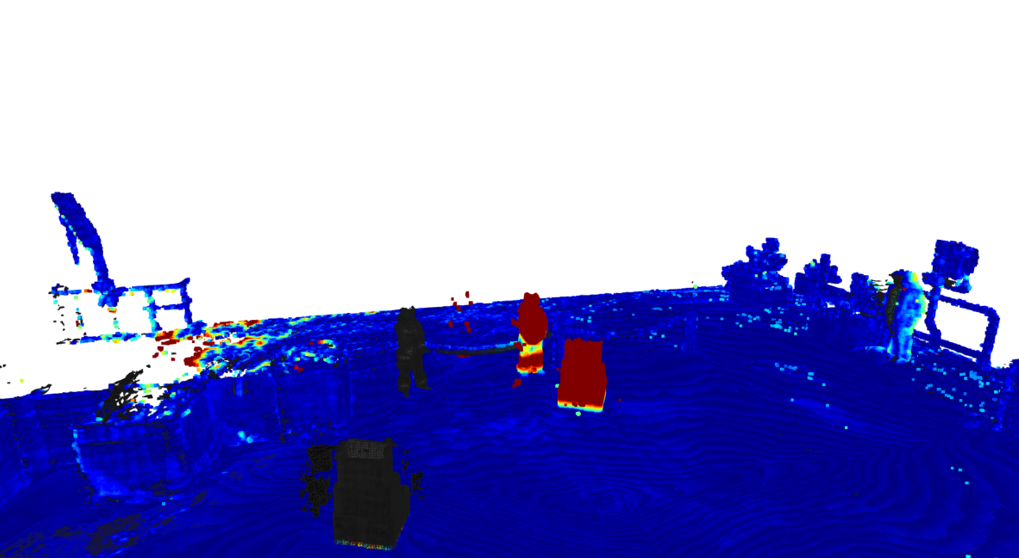}
\end{subfigure}
\caption{Top: Reference point cloud built by the Mapping LiDAR Setup. Colorization is based on surface reflectivity for infrared light. Bottom: Results of the change detection algorithm. In this experiment, objects and a person were relocated from one location (gray) to a new location (red). Colors represent the distance to the reference mesh.} \label{fig:mapping-lidar}
\end{figure}

Detecting and tracking geometric changes in industrial environments is crucial for ensuring safe plant operation. Moreover, it can be the keystone for predictive maintenance. To evaluate the capability of the module for 3D mapping and change detection multiple data recordings were obtained with the robot in the chemical plant Marl. The foundation is a highly detailed 3D point cloud captured by the Mapping LiDAR Setup as described in Section \ref{sec:mapping-lidar}. This reference model is converted into a continuous surface representation, a triangular mesh. Test point clouds recorded in other runs are then compared to this mesh by computing the nearest distances of all points to the reference mesh. In this way, different clusters of points can be identified that represent geometric change, as illustrated in Figure \ref{fig:mapping-lidar}. The proposed method allows for intuitive anomaly detection and assessment.

\subsection{Gas Detection Experiments}
\begin{figure}[htbp]
    \centering
    \includegraphics{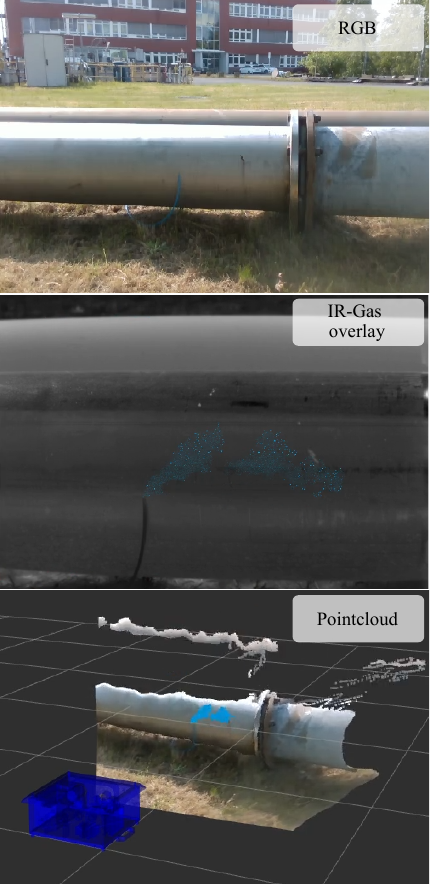}
    \caption{Different image streams of the active gas camera are shown filming an artificial methane leakage on an industrial plant. The registered point cloud (bottom) is possible through an RGB-Depth-IR calibration. A video of this scene, from which the images are taken, is published together with this work.}
    \label{fig:gas_camera_pipeline}
\end{figure}

The active gas camera was used to detect artificial gas leaks that we installed on an industrial site of a chemical park in Germany. Figure \ref{fig:gas_camera_pipeline} shows a $\sim$\SI{40}{\milli\liter\per\minute} methane leak that we place in front of a pipeline. The flow rate was set using a mechanical flowmeter from \textit{Brooks instruments} which had \SI{40}{\milli\liter\per\minute} as the lower limit. The leak was filmed from \SI{2}{\meter} distance. The point cloud is achieved by a geometric calibration of the stereo camera to the \ac{IR} camera. This allows to project the source of the gas leak to the depth frame and project it at the correct distance in the pointcloud. The gas overlay on the IR image is scaled such that stronger occupancy relates to a higher methane concentration length. The scene is also available as a video\footnote{\url{https://dx.doi.org/10.21227/4ms5-rs57}} published together with this article. The red-masked objects are detected as moving objects and are not considered to be gas information. A description of how these algorithms work can be found in \cite{bergau_real-time_2023}.

Methane leaks of $\sim$\SI{40}{\milli\liter\per\minute} could be visualized while this limit strongly depends on the wind conditions. It should be noted that the active approach is challenged by the wind as it uses subsequent images for the concentration calculation, in which wind disturbs the static scene. An improvement of the camera during this project resulted in stable methane detection up to wind speed of $\sim$\SI{2}{\meter\per\second} by minimizing the relevant times in between the recorded subsequent frames. Apart of hyperspectral imaging which is typically significantly more expensive, to the best of our knowledge, this is the first reported gas camera to visualize such small leaks under real-world conditions in real-time.

\subsection{Acoustic Experiments}
The acoustic experiments are structured into three distinct parts. Initially, a static phase is implemented to assess the \ac{DoA} estimation accuracy using the sound emitted by pumps. This phase serves to evaluate the precision of the \ac{DoA} estimator under controlled conditions.
Following the static phase, a dynamic experiment is executed, involving the robot being driven into the proximity of the pumps. The objective of this phase is to evaluate the source detection capabilities in a more dynamic setting, considering the robot's movement in relation to the pumps.
In the final part of the experiments, the focus shifts to anomaly detection. Various experiments are conducted to assess the system's ability to detect anomalies in the acoustic environment. This phase aims to evaluate the robustness and effectiveness of the anomaly detection mechanisms implemented in the system.

\begin{figure}
    \centering
    \input{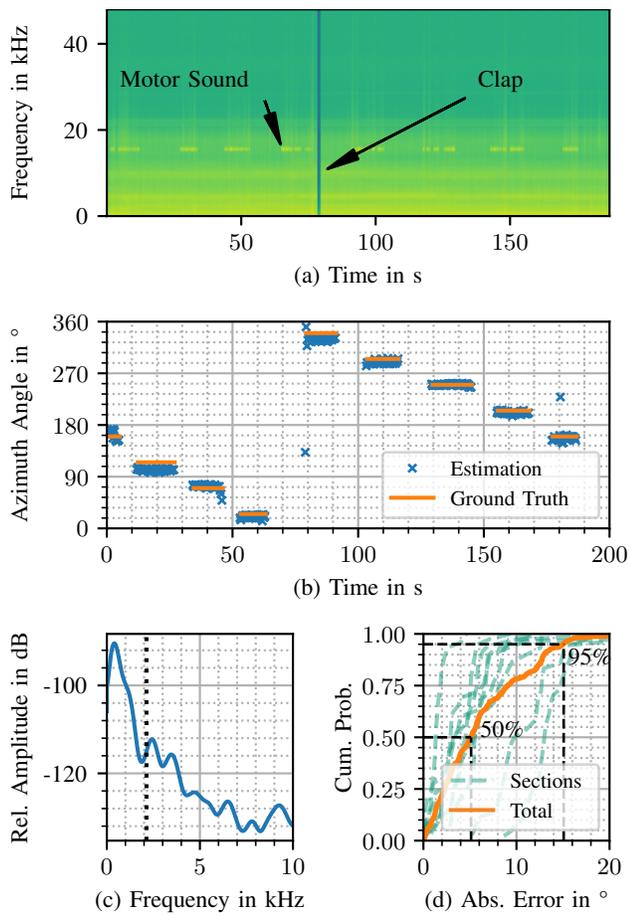}
	\caption{Static Experiment (a) Features the spectrogram of the recorded signal, highlighting the discernible motor noise just below 20 kHz during robot movement. A distinctive clap performed by the experimenter is denoted in the plot when the robot faces 180 degrees away from the working pump. (b) Displays the evolution of the estimated Direction of Arrival (DOA) throughout the experiment run. (c) Illustrates the frequency power distribution of the sounds emitted by the working pump. In this instance, a significant portion of the frequency content falls below the array's maximum spatial frequency. (d) Depicts the cumulative error distribution of the DOA error. The DOA estimation performance demonstrates a median circular error of approximately 5 degrees, with a 95\% confidence interval of around 15 degrees.}
	\label{fig:exp-acoustic-static}
\end{figure}

\paragraph{Static Experiments (Pump Equipment Localization)}
In this experiment, the robot is positioned in front of an operational pump. Subsequently, the robot undergoes rotational movement by a specific degree in each step, as illustrated in \autoref{fig:exp-acoustic-static}. The analysis reveals a median circular error of approximately \SI{5}{\degree}, accompanied by a 95\% confidence interval of around \SI{15}{\degree}.
The evaluation of the \ac{CDF} for individual sections indicates no significant deviation. Therefore, it can be inferred that the \ac{DoA} estimation performance remains isotropic across the azimuthal range.

\begin{figure*}[h!]
    \centering
    \input{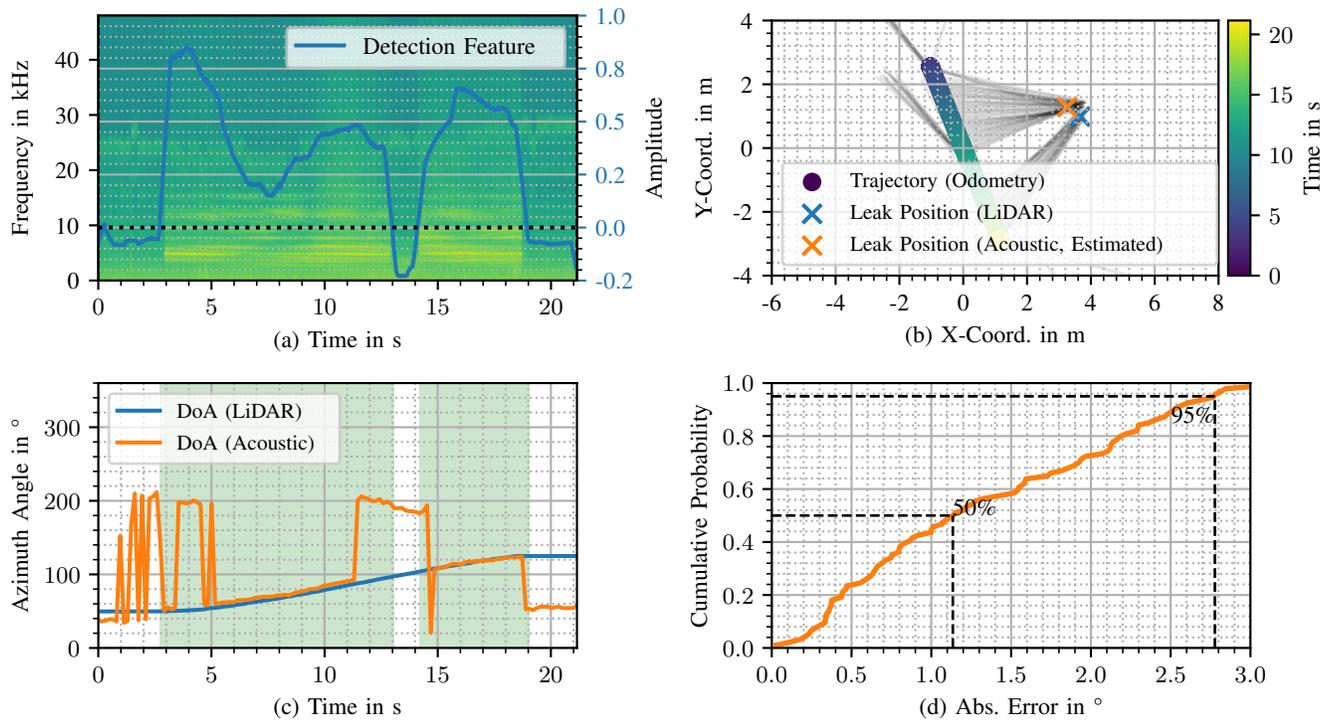}
	\caption{Dynamic Experiment (a) Depicts the spectrogram of the recording signal, incorporating the played-back sound of a leak. Overlayed on the spectrogram is the crafted detection feature designed for the leak signal. (b) Illustrates the trajectory of the robot during the experiment, with the measured leak positions obtained from both LiDAR and acoustic estimation. The single DoA measurements are represented by thin black lines. The position of the leak can be performed by acoustic means to an error of around \SI{0.5}{\meter} to the LiDAR measurement. (c) Presents a comparison between the estimated acoustic DOA values and the LiDAR-measured DoA values. Larger deviations happen when another sound source is interfering with the measurement. Shading highlights regions with a positive detection feature. (d) Displays the cumulative error distribution of the absolute circular error. The system demonstrates the capability to estimate with a median accuracy of \SI{1.13}{\degree} (or \SI{2.8}{\degree} at the 95th percentile).}
	\label{fig:exp-acoustic-dynamic}
\end{figure*}
\paragraph{Dynamic Experiment (Leak Localization)}
A sound speaker is positioned near a pipe, emitting a recorded sound mimicking a pipe leak. The robot follows a linear trajectory (refer to \autoref{fig:exp-acoustic-dynamic} (b)) during the experiment, and due to the recording's brevity, a brief interruption occurs. To create a detection feature, a simple inner product detection method is employed based on the frequency distribution of the leak signal.
Using the LiDAR sensor, the position of the speaker is established as a reference. Since a substantial portion of the leak signal surpasses the array's maximum spatial frequency, conventional DoA estimation methods discussed earlier are not applicable. Instead, the well-known \ac{SRP-PHAT} algorithm \cite{DiBiase.2000} is employed for DoA estimation in this experiment.
With the LiDAR reference, a DOA estimation accuracy of approximately \SI{1.13}{\degree} degrees (or \SI{2.8}{\degree}  degrees at the 95th percentile) is achieved, considering only the estimated values falling within regions where the detection feature exceeds 0. Combining the odometry-based position information from the robot with the estimated DOA values, the leak's position is determined.
This localization is executed by intersecting various pairs of generated lines, and the median of these intersections yields the leak's position. This estimated position is within \SI{0.5}{\meter} of the LiDAR measurement.

\paragraph{Anomaly Detection}



Two strategies were tested to detect sound anomalies, namely, reconstruction error, and latent space distance. 
In the reconstruction error strategy, the network tries to recreate the input sound signal using the intermediate latent vector. Since the network is solely trained on normal samples, it can recreate (decode) normal samples but fails to recreate an anomalous sound sample. The reconstruction error between the input and output samples is compared against a threshold value and used to differentiate between a normal and an anomalous sound sample. 
The latter strategy relies on the fact that latent vectors of similar objects are clustered near one another in the latent space, hence, the latent vector of an anomalous sample gets placed far from the latent vectors of the normal samples. An anomaly is detected when the distance between the latent vector of the current sample and the mean of all latent vectors from training is above a predefined threshold.


	The encoder-decoder model was initially trained on the MIMII\cite{purohit2019mimii} dataset, which comprises 
numerous sequences of normal and anomalous sound samples. From them, 4800 samples were used for training, 447 for validation, and 1524 for testing. The reconstruction error-based strategy achieved an accuracy of 69.62\% on the test set, while the latent distance-based approach achieved an accuracy of 82.28\% on the test set. 
Additionally, the model was further trained to include sound samples from the Marl chemical plant, adding splits of 32 train, 6 validation, and 7 test samples. With this extended dataset, the model achieved an accuracy of 84\% using the reconstruction error-based approach and 100\% using the latent distance-based approach. In the field, anomalies were simulated by playing "broken" sounds near the pumps.
\section{Conclusion and Outlook}
This paper introduced a robotic system designed for the comprehensive inspection of an industrial chemical plant. The system encompasses various key features, including navigation capabilities, and 2D object detection with a demonstrated general Average Precision (AP) of \num{0.7662}, covering objects such as cars and pedestrians. Through extrinsic calibration of multiple optical sensors, the system is proven to estimate the 3D position of detected objects, which, in turn, is employed for planning collision-free paths.
Additionally, the paper highlights the sensitivity and the real-world capability of the active gas camera. It visualizes an artificial methane leak of only \SI{40}{\milli\liter\per\minute} in real-time and successfully estimates its 3D position by combining depth information of the stereo camera up to \SI{2}{\meter} distance.
The final aspect presented is the elucidation of the working principles behind the acoustic pump equipment and leak localization. This section demonstrates the system's capability to localize a leak with an absolute error of around \SI{.5}{\meter} while the robot is in motion.

Routine inspection rounds are mandatory in many companies in process industries. The frequency and duration of these field rounds depend on the potential hazard of the plant, and they are conducted 24/7. The reduced availability of qualified candidates for such jobs, the desire to increase their appeal, and the need to collect reliable and objective sensor data fuel the development and testing of robots that can take over these tasks in diverse companies with industrial processes. Whereas navigation with legged robots in such plants seems already feasible, further development of adequate sensors and algorithms for the evaluation and interpretation of their signals is still necessary. Further, the ability of humans to depart from routine rounds when detecting an incident in peripheral vision is an open issue that needs to be transferred to such robots.

\bibliographystyle{IEEEtran}
\bibliography{references.bib}

\end{document}